\documentclass[11pt,twoside,a4paper]{article}
\usepackage[a4paper,top=35mm,bottom=35mm,left=35mm,right=35mm]{geometry}
\usepackage{xcolor}
\usepackage{times}
\usepackage[utf8]{inputenc}
\usepackage[T1]{fontenc}
\usepackage{hyperref}
\usepackage{graphicx}
\usepackage{booktabs}
\usepackage{academicons}
\usepackage{orcidlink}
\usepackage[misc]{ifsym}
\newcommand{\corr}{(\Letter)}

\usepackage{fancyhdr}
\usepackage{amsmath}
\usepackage{amsfonts}
\usepackage{amssymb}
\usepackage[ruled]{algorithm2e}
\usepackage{subcaption}
\usepackage{multirow}
\usepackage{makecell}
\usepackage{colortbl}
\usepackage{comment}
\usepackage[affil-it]{authblk}
\usepackage{etoolbox}
\DeclareMathOperator*{\argmax}{arg\,max}
\newcommand{\bftab}{\fontseries{b}\selectfont}

\title{Variance-Based Defense Against Blended Backdoor Attacks\thanks{This paper has been accepted at ECML PKDD 2025.}}

\author[1,3]{Sujeevan Aseervatham\orcidlink{0000-0001-8406-7795} \corr }
\author[1,2,3]{Achraf Kerzazi\orcidlink{0009-0001-1852-5927}}
\author[2,3]{Younès Bennani\orcidlink{0000-0003-3667-3357}}

\affil[1]{Orange Research, Châtillon, France \par \texttt{\small \{sujeevan.aseervatham, achraf.kerzazi\}@orange.com}}
\affil[2]{Universit\'e Sorbonne Paris Nord, LIPN, UMR 7030 CNRS \par \texttt{ \small younes.bennani@sorbonne-paris-nord.fr}}
\affil[3]{LaMSN - La Maison des Sciences Numériques, F-93210, Plaine Saint-Denis - France}

\date{}

\begin{document}

\maketitle

\begin{abstract}
Backdoor attacks represent a subtle yet effective class of cyberattacks targeting AI models, primarily due to their stealthy nature. The model behaves normally on clean data but exhibits malicious behavior only when the attacker embeds a specific trigger into the input. This attack is performed during the training phase, where the adversary corrupts a small subset of the training data by embedding a pattern and modifying the labels to a chosen target. The objective is to make the model associate the pattern with the target label while maintaining normal performance on unaltered data.

\sloppy
Several defense mechanisms have been proposed to sanitize training datasets. However, these methods often rely on the availability of a clean dataset to detect statistical anomalies, which may not always be feasible in real-world scenarios where datasets can be unavailable or compromised. To address this limitation, we propose a novel defense method that trains a model on the given dataset, detects poisoned classes, and extracts the critical part of the attack trigger before identifying the poisoned instances. This approach enhances explainability by explicitly revealing the harmful part of the trigger. The effectiveness of our method is demonstrated through experimental evaluations on well-known image datasets and a comparative analysis against three state-of-the-art algorithms: SCAn, ABL, and AGPD.
\end{abstract}
\begin{quote}
\end{quote}

\bigskip

\section{Introduction}
The rise of Artificial Intelligence (AI), and more specifically Deep Learning-based systems, has been dazzling and unparalleled. Its use is now widespread and, thanks to the availability of free software, a large public, even without specific knowledge, can build an AI model. They can also download training datasets and pre-trained models. Like any software, AI systems are vulnerable to malicious attacks. They can traditionally be attacked at the software layer, e.g., through vulnerabilities in the software libraries, but they can also be attacked through data and model-parameter manipulation. A corrupted AI can exhibit malicious behavior, which can have critical consequences. For example, self-driving cars may intentionally collide depending on the attacker's aim and a Large Language Model (LLM) may give fake news. This security risk has become a real-life threat and a major issue when researchers have shown that the AI models can be attacked in various ways depending on the aim, knowledge, and model access privilege of the attacker \cite{Survey2018,survey2019}. The attacker can lead an evasion attack by slightly modifying the model's input in order to change the decision/prediction. For example, for a spam recognition system, the attacker may change some words to bypass the filter. She can also lead inference attacks where the attacker wants to extract some information about the training dataset on which the model was trained. This attack is also known as privacy attack or model inversion attack. Indeed, in a personalized drug system, she may want to know if a particular patient information was in the training set. During the training stage, an attacker can also poison the training data to either degrade the model's performance (non-targeted attacks), e.g., by flipping the labels, or introduce a backdoor than can be exploited, during inference, to trigger a malicious behavior of the model. 

In this paper, we focus on a defense method against the backdoor attacks where the attacker wants a model to associate her data pattern to her target label. Thus, in the inference stage, any input containing the pattern will trigger the misclassification of the input as the target label. This is a vicious attack as the model seems to be pristine w.r.t. to its performance on clean data but in fact contains a backdoor which changes the prediction only in the presence of the trigger in the input data. For example, in a self-driving car, an attacker may have corrupted the model to recognize a green-square sticker pasted on a stop sign as a 70mph speed limit board. Such attacks can easily be performed when the attacker has a write access to the training data, as illustrated by the BadNets algorithm \cite{Gu2017}. More advanced attacks based on BadNets have been proposed, but they often require more resources or access rights.  By its simplicity, BadNets remains a threat that can be exploited by a large population, including opportunistic criminals and script-kiddies (open source script users without particular knowledge). Many defenses have been proposed in the literature, but most of them require an additional dataset of clean data on which statistics are computed to detect a distribution anomaly induced by a poisoned sample. In real-life, such dataset may hardly be available and even be subject to a corruption.

We propose a multi-stage defense algorithm that estimates the poisoned subset of the training data, detects the poisoned classes, extracts the attack trigger for each class, and identifies the poisoned instances based on the extracted pattern.
Our contributions can be summarized in three points: 1) we introduce a novel defense algorithm against BadNets \cite{Gu2017} and Blended \cite{blended2017} attacks which does not rely on additional prerequisites such as the availability of a clean dataset; 2) the defense is explainable, as it extracts the most important part of the trigger responsible for the malicious behavior; and 3) our method is effective in All-to-All attacks where many classes are poisoned. 

The remainder of this paper is organized as follows: in Section~\ref{sec:related_word}, we present the state-of-the-art of backdoor attacks and defenses. Our defense algorithm is detailed in 
Section~\ref{sec:vbd}. Section~\ref{sec:experiments} describes the experimental results and, finally, Section~\ref{sec:conclusion} concludes this article.

\section{Related Work} \label{sec:related_word}
\subsection{Backdoor Attacks}
The Backdoor attack was introduced in \cite{Gu2017} where the proposed BadNets methodology involves patching a small pattern onto the input data of a small percentage of the training set and labeling them with the desired target label. By training on this poisoned set, the model learns to associate the attack pattern with the target label. Thus, the presence of the pattern in an input instance triggers the malicious behavior of the model by classifying the input as the attacker's target label instead of the correct one. To make the pattern more stealthy, linear blending is used in \cite{blended2017}, while in \cite{lowfreq2021} the pattern is generated in the low-frequency domain by considering both the training set and a pre-trained model. These works paved the way for more advanced attacks, where the pattern is not fixed but specific to the input \cite{nguyen2020input,nguyen2021wanet}. However, compared to fixed pattern attacks, input-based pattern attacks require more prerequisites, such as access to and modification of the training algorithm, making them more suitable when targeting users who download pretrained models. Fixed-pattern attacks remain more realistic in real-world scenarios, especially when the attacker is an insider with access to the training data.

\subsection{Backdoor Defenses}
In the last few years, many backdoor defense methods have been proposed by researchers, each designed to be used at a specific stage of the AI model lifecycle. As noted in \cite{activationgradient2024} and \cite{zhang2024reliable}, these defenses can be grouped in four categories based on the lifecycle stage: 1) pre-training stage methods, mainly used to detect the poisoned instances and sanitize the dataset \cite{AC2019,strip2019,chou2020sentinet,ScanTang2021demon,Beatrix2023,latentSep2023,activationgradient2024}, 2) in-training stage defenses, which aim to reduce the effect of the poisoned data on the model \cite{li2021anti,DecouplingTraining2022,zhang2024reliable}, 3) post-training stage methods, used to correct backdoored models \cite{advneurpruning2021,NeuronPruning2023}, and 4) inference stage defenses,  which are used to detect malicious inputs by analyzing both both the input and the output of the model, typically residing between the user and the model \cite{scaleup2023}. These methods can also be categorized based on the information they use to mitigate the attacks. The main approaches are input-based, loss-based, and activation-based. Defenses in the pre-training and inference stages are mainly input-based or activation-based methods, while algorithms in the in-training and post-training stages are mostly loss-based or activation-based. Input-based approaches operate on the input data by altering it and computing statistical measures on the predictions, such as the entropy  \cite{strip2019,chou2020sentinet,scaleup2023}. In \cite{strip2019}, the STRIP algorithm linearly blends the input image with a set of clean images before computing the entropy of the prediction on each perturbed images. A poisoned image is detected when the entropy is low. In \cite{chou2020sentinet}, the SentiNet method uses Grad-CAM \cite{gradcam} to extract the decision region from an input image and superimposes it on a set of clean images to check whether it triggers a misclassification of the model.
Activation-based methods rely on the assumption that the poisoned and clean samples can be separated, e.g., through clustering, in the feature space induced by the activations of a specific layer of the model \cite{AC2019,ScanTang2021demon,Beatrix2023,latentSep2023}. In \cite{AC2019}, K-Means clustering is used on the activations and in \cite{ScanTang2021demon}, the proposed SCAn algorithm uses a two-component Gaussian Mixture Models (GMM) for clustering. In the loss-based approach, the methods  rely on the property that poisoned instances are classified with a very low value of the loss. In \cite{li2021anti}, the Anti-Backdoor Learning (ABL) defense uses the loss to isolate the poisoned instances during the training and then unlearns the poisoned data through gradient ascent.

\section{Variance-based Defense} \label{sec:vbd}
\subsection{Threat Model}
\sloppy
In this paper, we assume that the attacker has full access to the training database. The attacker aims to modify the model's behavior so that any input patched with her attack pattern is classified as her target class, while maintaining good categorization performance on pristine inputs to remain undetected.  To implement this attack, she employs a combination of BadNets~\cite{Gu2017} and Blended~\cite{Chen2017} attacks. Given an attack pattern/trigger ${\bf p}$, its corresponding binary mask ${\bf m}$, the target label $y_t \in \mathcal{Y}$, and a blending factor $\alpha \in [0,1]$, she selects a subset of the training dataset with a ratio $r$ and generates malicious data according to Equation~\ref{eq:transf}, before adding it to the training set with her target label. The poisoned dataset is then distributed or made available to download. Training a model on this poisoned dataset will cause the model to capture a relationship between the attack trigger ${\bf p}$ and the target label $y_t$. 

\begin{equation}
\tilde{{\bf x}} = \Gamma({\bf x}, {\bf p}, {\bf m}, \alpha) = {\bf x} \odot ({\mathbf{1} } - {\bf m}) + (1-\alpha)\cdot({\bf x} \odot {\bf m}) + \alpha \cdot ({\bf m} \odot {\bf p})
\label{eq:transf}
\end{equation}
where ${\bf x} \in \mathcal{X}$, with $\mathcal{X}\subseteq \mathbb{R}^{H\times W \times C}$, is an input image of height $H$, width $W$ and color dimension $C$, $\odot$ the element-wise tensor product operator and $\mathbf{1}$ the all-ones tensor of the same dimension as $m$.

\subsection{Motivation}
Before using a dataset to train a model, it is important to sanitize it and ensure that no malicious data leading to a blended or a BadNets backdoor are present. To achieve this, we propose an algorithm to detect and extract the trigger pattern from the training dataset. Using the extracted patterns, the dataset can be sanitized by removing the data containing these patterns.\\
We want to extract the pattern ${\bf p_t}$ and its binary mask ${\bf m_t}$ for a target label $y_t$ based on the following hypotheses:
\begin{enumerate}
\item when a pristine input is patched with the pattern ${\bf p_t}$, the model must predict $y_t$, which means that ${\bf p_t}$ and ${\bf m_t}$ should minimize the training loss for the set of clean data ($\mathcal{D}_{C}$) patched with ${\bf p_t}$ and associated with the target label $y_t$,
\item the pattern should also be as small as possible, i.e., only a few elements of the binary mask should be set to 1 (this is equivalent to using a $L_0$ norm penalization on ${\bf p_t}$) in order to keep the malicious data unnoticeable among the training instances,
\item the pixels of the pattern should be located at low-variance positions in a variance image computed from a set of malicious data with the label $y_t$ ($\mathcal{D}_{P_t}$), since we are defending against a static-trigger data poisoning.
\end{enumerate}

Given these three hypotheses, we can formulate the following minimization problem to compute $({\bf p_t},{\bf m_t})$:
\begin{equation}
\min_{p,m}  \frac{1}{|\mathcal{D}_{C}|}\sum_{({\bf x}^{(k)}, y^{(k)})\in\mathcal{D}_{C}} \ell(f(\Gamma({\bf x}^{(k)}, {\bf p}, {\bf m}, \alpha)), y_t) + \lambda \cdot \lVert {\bf m} \rVert_{1} + \gamma \cdot \lVert {\bf m} \odot V_{\mathcal{D}_{P_t}} \rVert_{1}
\label{eq:var_min}
\end{equation}
where $f$ is the poisoned model learned from the malicious dataset, $\mathcal{D}_{C}$ the clean part of the training set, $\mathcal{D}_{P_t}$ the poisoned part with label $y_t$, $\ell$ the loss function, usually the Cross-Entropy loss, $\lambda$  and $\gamma$ the loss terms weighting factors and $V_{\mathcal{D}_{P_t}}$ the mean, over the color channel, of the min-max-scaled variance of the data in $\mathcal{D}_{P_t}$ such that:

\begin{equation}
[V_{\mathcal{D}_{P_t}}]_{i,j} = \frac{1}{n_{r}}{\sum_{c\in\{1,\ldots,n_r\}} \frac{\text{var}(X_{i,j,c}; \mathcal{D}_{P_t}) - \min_{k,l}(\text{var}(X_{k,l,c}; \mathcal{D}_{P_t}))}{\max_{k,l}(\text{var}(X_{k,l,c} ; \mathcal{D}_{P_t})) - \min_{k,l}(\text{var}(X_{k,l,c} ;\mathcal{D}_{P_t}))}}
\label{eq:scaledvariance}
\end{equation}
with $X_{i,j,c}$ the random variable associated with the pixel/variable $x_{i,j,c}$ where $c$ is the color channel index, $n_r$ the channel dimension and the variance given by: 
\begin{equation}
\text{var}(X_{k,l,c} ;\mathcal{D}_{P_t}) = 
\frac{1}{|\mathcal{D}_{P_t}|}\sum_{({\bf x}^{(n)}, y^{(n)})\in\mathcal{D}_{P_t}} ({\bf x}^{(n)}_{k,l,c} - \overline{X_{k,l,c}})^2
\end{equation}

Solving the problem~\ref{eq:var_min} may be time-consuming and difficult since $\mathcal{D}_{C}$ and $\mathcal{D}_{P_t}$ are not known, and it may even lead to adversarial noises, especially if the model is not robust.\\
Instead of solving directly this minimization problem, we propose in the next section a heuristic to approximate the patterns.

\subsection{Method}

\begin{figure}[!htb]
\centering
\includegraphics[width=12cm,angle=0]{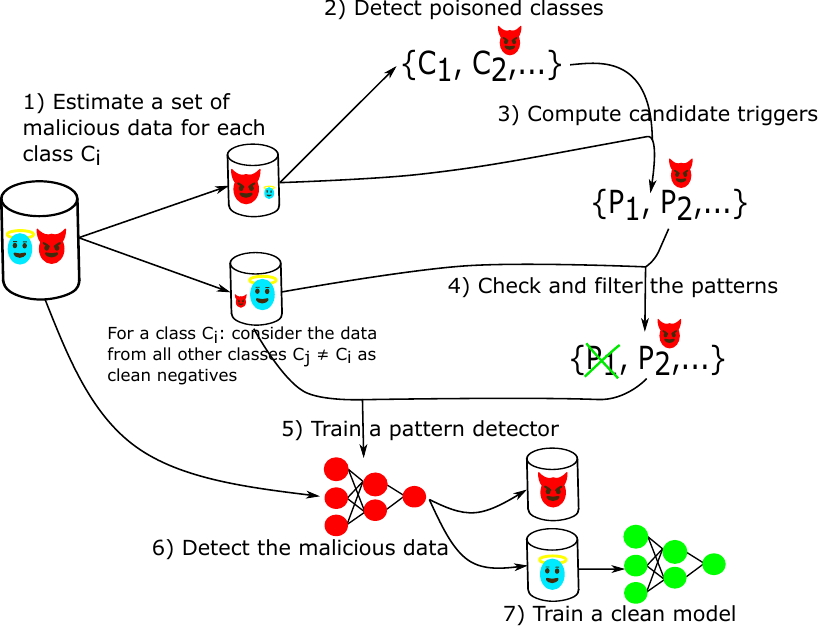}
\caption{Workflow of the proposed sanitization method.} 
\label{fig:workflow}
\end{figure}

The proposed data sanitization method is described in Algorithm~\ref{alg:vbd} and illustrated step by step in Figure~\ref{fig:workflow}. It consists of seven main steps, listed as follows:

\begin{enumerate}
\item Estimate a set of malicious instances ${\tilde{\mathcal{D}_{P}}}$
\item Detect the potentially poisoned classes
\item Compute a candidate pattern for each poisoned class
\item Check the patterns on an estimated ${\tilde{\mathcal{D}_{C}}}$
\item Train a pattern detector for each detected pattern
\item Detect the malicious instances
\item Train a clean model
\end{enumerate}

\begin{algorithm}
\caption{Variance-based defense}\label{alg:vbd}
\KwData{$\mathcal{D}$: the training set to sanitize}
\KwResult{$\mathcal{D}_{P}$: the set of malicious instances and $f_c$: the sanitized model}
\ForEach{class $c_i \in \mathcal{Y}$}{
${\tilde{\mathcal{D}_{P_i}}} \gets $ Estimate a set of malicious instances with label $c_i$ \;
\If {$c_i$ \text{is detected as poisoned}}{
$({\bf p_i}, {\bf m_i}) \gets$ Compute a candidate pattern and its mask based on $\tilde{\mathcal{D}_{P_i}}$ \;
}
}
$\mathcal{P}\gets$ Check and filter the candidate pattern set $\{({\bf p_i}, {\bf m_i})\}_{i}$ \;
\ForEach{$({\bf p_i}, {\bf m_i}) \in \mathcal{P}$}{
$h_i \gets $ Train a model to detect the instances patched with ${\bf p_i}$ \;
$\mathcal{D}_{P_i} \gets $ Use $h_i$ to identify the malicious instances in $\mathcal{D}$ \;
}
$\mathcal{D}_{P} \gets \bigcup_i \mathcal{D}_{P_i}$ \;
$f_c \gets$ Train a clean model using $\mathcal{D}$ and $\mathcal{D}_{P}$ \;
\end{algorithm}

\noindent In the following paragraphs, we provide a detailed description of each step.

\subsubsection{Stage 1: Malicious instance set estimation}
\begin{figure}[!htb]
\centering
\includegraphics[width=4cm,angle=90]{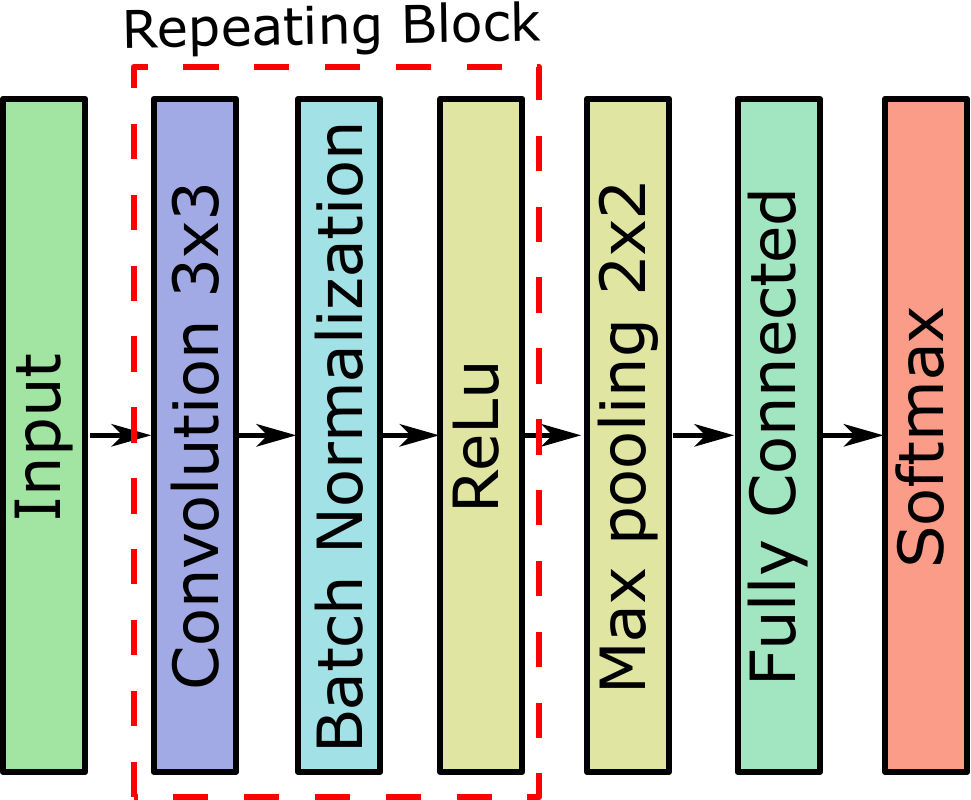}
\caption{Architecture of the classifier used to overfit the attack patterns.} \label{fig:simpleclassifier}
\end{figure}

We assume that the patterns used to poison the training set are small and simple, such that a simple classifier with a single convolution layer can memorize the relation between the pattern and the target label. The idea is to train a simple convolution model with a few layers as the one shown in Figure~\ref{fig:simpleclassifier} on the training set to overfit the attack pattern. The model is thus expected to perform well on the poisoned instances and have a high generalization error on the clean data. We refer to this model as the {\it simple poisoned model}, denoted $f_s$. A subset of the poisoned data, for a class $c_i$, can then be estimated by selecting the top $N$ instances with the label $c_i$ and correctly predicted with the highest probability score ($N$ being a parameter, knowing that for our experiments we set $N=20$). 

\subsubsection{Stage 2: Poisoned classes detection}
At this stage, we aim to filter out the classes considered as non-poisoned, retaining only the suspicious ones for further analysis. To achieve this, for each class $c_i$ we identify the most important (pixel) variable that is commonly used by the poisoned model $f_s$ to correctly classify the instances of the estimated malicious set $\tilde{\mathcal{D}_{P_i}}$. If the empirical distributions of the importance of this variable in $\tilde{\mathcal{D}_{P_i}}$ and in a subset of potential clean-instances from $c_i$ are the same then we can assume that the class $c_i$ is not poisoned. To find the most important variable, given the {\it simple poisoned model} $f_s$, we compute the gradient of the first loss term of Equation~\ref{eq:var_min} w.r.t. the malicious input, as given by Equation~\ref{eq:patgrad}. 

\begin{equation}
\nabla L(f_s,\tilde{\mathcal{D}_{P_i}}) = \frac{1}{|\tilde{\mathcal{D}_{P_i}}|}\sum_{({\bf x}^{(k)}, c_i)\in\tilde{\mathcal{D}_{P_i}}} \frac{\partial \ell}{\partial {\bf x}}(f_s({\bf x}^{(k)}), c_i)
\label{eq:patgrad}
\end{equation}
As we want the most important pixels explaining the decision, we choose the following loss function $\ell({\bf x},c)$:
\begin{equation}
\label{eq:losspatgrad}
\ell({\bf x},c) = P(c|{\bf x};f_s) - \max_{y \in{\mathcal{Y}-\{c\}}} P(y|{\bf x};f_s)
\end{equation}
$P(c|{\bf x};f_s)$ is given by the $c^{\text th}$ component of the softmax layer of $f_s$ and in order to avoid numerical instabilities, we use the logit layer instead of the softmax layer.\\
~\\
For poisoned instances, some components of the gradient (Eq.~\ref{eq:patgrad}) will have a large absolute value compared to the clean-instances. Indeed, the partial derivative at given coordinates indicates the sensitivity of the loss function to changes in that pixel's value. In the case of poisoned instances, the backdoor pattern is designed to strongly influence the model's prediction. As a result, only a few pixels, those containing the trigger, tend to dominate the gradient signal. We thus assume that, for poisoned instances, a small number of pixels will have a  large impact on the prediction. The most influential pixel can therefore be identified by locating the index $(i^*,j^*)$ corresponding to the maximum absolute value of the mean gradient across the color channels:
\begin{equation}
i^*, j^* = \argmax_{i, j} ~ \lvert \text{mean}_c \left[\nabla L(f_s,\tilde{\mathcal{D}_{P_i}})\right]_{i,j,c} \rvert 
\end{equation}
We now define the importance of this pixel in the decision of an input ${\bf x}$ by $f_s$, as follows:
\begin{equation}
I({\bf x},c_i; f_s, i^*,j^*) = \max_c \left[ \lvert \nabla L(f_s,\{({\bf x},c_i)\})\rvert \right]_{i^*,j^*,c}
\end{equation}
~\\
The two-sided Kolmogorov–Smirnov test \cite{ks2008} is used with a confidence level of 99\% to compare the distributions of $I({\bf x},c_i; f_s, i^*,j^*)$ between $\tilde{\mathcal{D}_{P_i}}$ and a set of potential clean-instances from $c_i$. We use the instances of $c_i$ with the lowest prediction scores as clean-instances. When the p-value is below $0.01$, we consider $c_i$ as potentially poisoned.

\subsubsection{Stage 3: Candidate pattern computation}
With the {\it simple poisoned model} $f_s$ and the estimated subset of malicious instances of the class $c_i$, we can compute an approximate pattern by using the gradient defined in Equation~\ref{eq:patgrad} which gives the importance of the pixels w.r.t. the loss function. To obtain a binary mask from the gradient, the following processing is performed:
\begin{enumerate}
\item Flattening the gradient of Eq.~\ref{eq:patgrad} by taking the $L_2$ norm of the gradient over the color channel
\item Min-Max-scaling the result to have the values within $[0, 1]$
\item Binarizing the values using a threshold
\end{enumerate}
The binarization step may be tricky as it involves defining a threshold which may depends on the input data. To avoid manually defining this threshold, we used the Otsu method, which finds the best threshold that minimizes the intra-variance \cite{Otsu1979}. In order to capture also the neighboring points of an important point, we apply a Gaussian blur before applying the Otsu method.\\
Once we have the binary gradient mask, we need to select only the points with the lowest variance, which requires a variance threshold. The variance matrix on $\tilde{\mathcal{D}_{P_i}}$ can be computed with the Equation~\ref{eq:scaledvariance}. The final candidate mask is obtained by performing an element-wise product between the gradient mask and the variance matrix before binarizing with the Otsu method (note that to keep the points with low variances, we use 1 minus the variance matrix).\\
To have the candidate pattern, we apply the candidate mask on the mean image of $\tilde{\mathcal{D}_{P_i}}$.
Algorithm~\ref{alg:patcomp} describes the whole process of generating the candidate pattern and mask.

\begin{algorithm}
\caption{Pattern Computation}\label{alg:patcomp}
\KwData{$\tilde{\mathcal{D}_{P_i}}$: a set of malicious instances for the class $c_i$, $f_s$: the poisoned model}
\KwResult{$({\bf p_i}, {\bf m_i}))$: the candidate pattern and mask for the class $c_i$}
${\bf g} \gets $ compute the gradient with Equation~\ref{eq:patgrad} using the logit layer \;
\ForEach{pixel at coordinate $i,j$} {
$\tilde{g}_{i,j} \gets \sqrt{\sum_r g_{i,j,r}^2}$ \;
}
${\bf g}_{\text scaled} \gets$ min\_max\_scale($\tilde{\bf g}$) \;
${\bf g}_{\text blured} \gets$ gaussian\_blur(${\bf g}_{\text scaled}$) \;
${\bf g}_m \gets$ otsu\_binarization(${\bf g}_{\text blured}$) \;
${\bf V}_{\tilde{\mathcal{D}_{P_i}}} \gets $ Use Equation~\ref{eq:scaledvariance} to compute the variance on $\tilde{\mathcal{D}_{P_i}}$\;
${\bf m_i} \gets \text{otsu\_binarization}({\bf g}_m \odot ({\mathbf{1} } -V_{\tilde{\mathcal{D}_{P_i}}})) $ \;
${\bf p_i} \gets {\bf m_i} \odot \frac{1}{|\tilde{\mathcal{D}_{P_i}}|}\sum_{({\bf x}^{(k)}, y^{(k)}) \in \tilde{\mathcal{D}_{P_i}}} {\bf x}^{(k)}$ \;
\end{algorithm}

\subsubsection{Stage 4: Check and filter the patterns}
At this stage, a candidate pattern and its mask have been calculated for each potentially poisoned class. We need to check that it triggers a malicious behavior of the {\it simple poisoned model}. To this end, we patch the pattern to instances from other classes (estimated ${\tilde{\mathcal{D}_{C}}}$) before feeding them to the model, and we check whether the model assigns them to the class associated with the pattern. We compute the Attack Success Rate (ASR), i.e., the proportion of patched instances predicted as the attack label. If this rate falls below a user-defined threshold, we remove the pattern from the candidate set. The remaining patterns are then considered harmful.

\begin{figure}[t]
\centering
\begin{subfigure}{.25\textwidth}
  \centering
  \includegraphics[width=2.1cm]{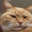}
  \caption{input}
  \label{fig:pattcompsfiga}
\end{subfigure}%
\begin{subfigure}{.25\textwidth}
  \centering
  \includegraphics[width=2.1cm]{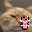}
  \caption{malicious}
  \label{fig:pattcompsfigb}
\end{subfigure}%
\begin{subfigure}{.25\textwidth}
  \centering
  \includegraphics[width=2.1cm]{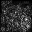}
  \caption{gradient}
  \label{fig:pattcompsfigc}
\end{subfigure}%
\begin{subfigure}{.25\textwidth}
  \centering\hspace{2.1cm}
\end{subfigure}%

\begin{subfigure}{.25\textwidth}
  \centering
  \includegraphics[width=2.1cm]{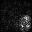}
  \caption{mean gradient}
  \label{fig:pattcompsfigd}
\end{subfigure}%
\begin{subfigure}{.25\textwidth}
  \centering
  \includegraphics[width=2.1cm]{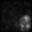}
  \caption{blurred}
  \label{fig:pattcompsfige}
\end{subfigure}%
\begin{subfigure}{.25\textwidth}
  \centering
  \includegraphics[width=2.1cm]{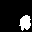}
  \caption{binarized}
  \label{fig:pattcompsfigf}
\end{subfigure}%
\begin{subfigure}{.25\textwidth}
  \centering\hspace{2.1cm}
\end{subfigure}%

\begin{subfigure}{.25\textwidth}
  \centering
  \includegraphics[width=2.1cm]{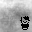}
  \caption{Variance}
  \label{fig:pattcompsfigg}
\end{subfigure}%
\begin{subfigure}{.25\textwidth}
  \centering
  \includegraphics[width=2.1cm]{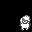}
  \caption{grad$\odot$(1-var)}
  \label{fig:pattcompsfigh}
\end{subfigure}%
\begin{subfigure}{.25\textwidth}
  \centering
  \includegraphics[width=2.1cm]{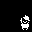}
  \caption{re-binarized}
  \label{fig:pattcompsfigi}
\end{subfigure}%
\begin{subfigure}{.25\textwidth}
  \centering
  \includegraphics[width=2.1cm]{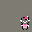}
  \caption{Extracted pattern}
  \label{fig:pattcompsfigj}
\end{subfigure}%

\caption{Illustration of the candidate pattern computation process. The first row shows (\subref{fig:pattcompsfiga}) an example of input image from the cat class, (\subref{fig:pattcompsfigb}) its malicious version labeled with the attacker's label  and (\subref{fig:pattcompsfigc}) its gradient (Eq.~\ref{eq:patgrad}). In the last two rows,  the pattern computation process, described in the Algorithm~\ref{alg:patcomp}, is illustrated step by step. The gray background in (\subref{fig:pattcompsfigj}) represents transparency.}
\label{fig:pattcomp}
\end{figure}

\subsubsection{Stage 5: Train a pattern detector} To detect the presence of a pattern in the training set, we propose to train a binary classifier for each pattern. We use a one-layer-convolution network as shown in Figure~\ref{fig:simpleclassifier} except that 1) we remove the batch normalization, 2) we add a dropout layer before the fully connected layer in order to reduce the overfitting and 3) we replace the softmax layer by a sigmoid layer. For a given pattern ${\bf p_i}$ for the class $c_i$, we build the training data as follows, we use the instances of the other classes and label them as "0" (clean), and we patch these instances using the Equation~\ref{eq:transf} with random blending factors, before labeling them as "1" (poisoned). Instead of training directly on the images, we can achieve better performance by training the model on the image gradient using Equation~\ref{eq:patgrad} on a single image. Moreover, we use Semi-Supervised Learning (SSL) with Pseudo-labels \cite{Lee2013PseudoLabelT}. After a few epochs of supervised learning, we use the classifier to label the data of the class $c_i$ and we add these data to the training set. We continue the SSL by relabeling the data after each epoch for a predefined number of epochs to make the pseudo-labels stable.
The Algorithm~\ref{alg:patdetect} describes the training procedure.

\begin{algorithm}
\caption{Training a Pattern Detector}\label{alg:patdetect}
\KwData{$\mathcal{D}$: the (poisoned) training set, $f_s$ the poisoned model learned in the first stage, $({\bf p_i}, {\bf m_i}, c_i)$ the attack pattern ${\bf p_i}$ with its mask ${\bf m_i}$ for the target class $c_i$ and $h_i$ the binary classifier to fit}
\KwResult{$h_i$: the trained pattern detector for $({\bf p_i}, {\bf m_i}, c_i)$}
$\mathcal{D}_T \gets \mathcal{D} - \{({\bf x}, y) \in \mathcal{D} : y \neq c_i \}$ \;
$\mathcal{D}_{ssl} \gets \{\} $ \;
$(\mathcal{D}_t, \mathcal{D}_v, \mathcal{D}_e)  \gets$ split $\mathcal{D_T}$ into training, validation, and test sets\;
\For{epoch $\gets0$ \KwTo max\_epoch}{
\ForEach{\text{batch} $\mathcal{B}$ of $\mathcal{D}_t$}{
    $\mathcal{U} \gets \{\}$ \;
    \ForEach{$(x, y) \in \mathcal{B}$}{
    $\alpha \gets$ random$(0.1, 1)$\;
    ${\bf z} \gets \Gamma({\bf x}, {\bf p_i}, {\bf m_i}, \alpha)$ (Eq.~\ref{eq:transf})\;
    $\mathcal{U} \gets \mathcal{U} \cup \{(\nabla L(f_s, \{ ({\bf x}, c_i)\}), 0), (\nabla L(f_s, \{({\bf z}, c_i)\}), 1)\}$ (Eq.~\ref{eq:patgrad}) \;
    }
    Optimize $h_i$ with $\mathcal{U} \cup \mathcal{D}_{ssl}$ \;
}
\If{ {start\_ssl\_epoch} $<$ epoch $<$ {stop\_relabeling\_epoch} }{
    Use $h_i$ to label $\{({\bf x}, y) \in \mathcal{D} : y = c_i \}$ \;
    $\mathcal{D}_{ssl} \gets $ the pseudo-labeled data \;
}
}
\end{algorithm}

\subsubsection{Stage 6: Detect the malicious instances} We compute the gradient for each instance of the training set, and we feed it to each trained pattern detector. We label the instance as malicious if at least one pattern detector labeled it as "1".

\subsubsection{Stage 7: Train a clean model} Once we have identified the malicious instances, we can train a clean model from scratch or fine-tune a poisoned model by using both the clean instance set and the poisoned one.

\subsubsection{Robustness} The proposed method relies on a {\it simple poisoned model}, which is assumed to separate poisoned data from clean data based on the prediction score. This implies that the model must be complex enough to capture the trigger (high attack success rate) and simple enough to underperform on clean data (low accuracy rate). Relying on a single model to achieve this task might not be robust. To address this issue, we propose to use an ensemble of simple models based on the architecture shown in  Figure~\ref{fig:simpleclassifier} with, e.g., different number of filters and layers. We use each model of the ensemble, simultaneously and independently, to perform Stage 1 to Stage 4. At the end of Stage 4, for each model, we have extracted a set of triggers, one for each detected poisoned class. For robustness, a class is considered poisoned only if the number of extracted patterns for that class exceeds a certain threshold, e.g., half the number of models in the ensemble (majority vote). However, in this work, we adopt a more defensive approach to reduce the false negative rate. Therefore, we consider a class as poisoned if at least one trigger has been extracted for this class. To perform the Stages 5 and 6, only one model and pattern must be retained for a given poisoned class. To select the most appropriate model-and-pattern pair for a class, we choose the one with the lowest accuracy on the training set (computed in Stage 1 during the training) and the highest ASR (computed in stage 4), i.e., the pair that achieves the highest score according to the following metric: $\alpha\cdot (1-\text{acc})+(1-\alpha)\cdot \text{asr}$. In our work, we set $\alpha=0.6$ to slightly favor the model with the lowest accuracy.

\section{Experiments} \label{sec:experiments}
\subsection{Experimental Setup}

For the evaluation, we used the BackdoorBench framework \cite{backdoorbench}, and our code is available online\footnote{\url{https://github.com/Orange-OpenSource/BackdoorBench/tree/vbd-v1}}. 
We compared our method, {\it VBD}, against three high-performing state-of-the-art methods \cite{activationgradient2024}: Anti-Backdoor Learning (ABL) \cite{li2021anti}, AGPD \cite{activationgradient2024}, and SCAn \cite{ScanTang2021demon}. For VBD, we used an ensemble of 3 models: a 1-layer model with 64 filters, a 2-layers model with 10 filters each, and a 2-layers model with 16 filters each, following the architecture shown in Figure~\ref{fig:simpleclassifier}. We used 80\% of the training set only to train the models and the remaining 20\% for the validation, detection, pattern extraction, and detector training. For training the poisoned and detector models, we used a learning rate of 0.01, a batch size of 256 for CIFAR-10 and 64 for Tiny ImageNet, and 20 training epochs. For the competitors, we used the default settings, which include the use of the Pre-Act ResNet18 model \cite{he2016identity}, knowing that AGPD and SCAn require also an additional dataset of clean images, which, by default, consists of 10 images per class taken from the test set. Full implementation details and additional parameters are available in the released code repository.\\
The experiments consist of poisoning 10\% of the training dataset using the BadNets and Blended attacks and evaluating the performance of the four methods in detecting the poisons in the training set. For the Blended attacks, we evaluated 3 blending factors: 10\%, 20\% and 50\%. A blending factor of 10\% means that the trigger is 90\% transparent. A BadNets attack is equivalent to a Blended attack with a blending factor of 100\% (full opacity). We used both an All-to-One attack, where only one class is poisoned with 10\% of the images from each class (including the target class) and an All-to-All attack, where all classes are poisoned such that class ($k+1 ~\text{mod}~ N_c$), with $N_c$ being the number of classes in the dataset, is poisoned with 10\% of the images from class $k$.

For the database, we used both CIFAR-10 \cite{Krizhevsky09} and Tiny ImageNet \cite{le2015tiny}. CIFAR-10 is composed of 10 classes with 5000 training images of size $32\times 32$ per class. Tiny ImageNet contains 200 classes with 500 training images of size $64\times 64$ per class. For AGPD and SCAn, which require 10 auxiliary clean images per class in their default settings, we used a total of 100 and 2000 clean images from the test set for CIFAR-10 and Tiny ImageNet, respectively.

\begin{figure}[t]
\hspace{.2\textwidth}
\begin{subfigure}{.2\textwidth}
\centering
  \includegraphics[width=1cm]{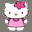}
  \hspace*{0.70cm}
  \caption{Big Kitty\hspace*{0.60cm}}
\end{subfigure}\hspace{0.3cm}
\begin{subfigure}{.2\textwidth}
  \centering
  \includegraphics[width=1cm]{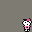}
  \hspace*{0.70cm}
  \subcaption{Small Kitty}
\end{subfigure} \hspace{0.12cm}
\begin{subfigure}{.2\textwidth}
  \centering
  \hspace*{0.03cm}
  \includegraphics[width=1.85cm]{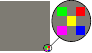}
  \caption{Colored grid}
\end{subfigure}

\hspace{.2\textwidth}
\begin{subfigure}{.2\textwidth}
  \centering
  \includegraphics[width=1.85cm]{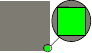}
  \caption{Green square}
\end{subfigure}\hspace{0.3cm}
\begin{subfigure}{.2\textwidth}
  \centering
  \includegraphics[width=1.85cm]{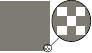}
  \caption{White grid}
\end{subfigure}\hspace{0.3cm}
\begin{subfigure}{.2\textwidth}
  \centering
  \includegraphics[width=1.85cm]{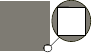}
  \caption{White square}
\end{subfigure}%
\caption{Attack patterns used in the evaluation. All the triggers, except for the two "kitty" patterns, are of size $3\times 3$. The "Big Kitty" spans the entire image and the "Small Kitty" has a size of $10\times 14$. The gray background is transparent. }
\label{fig:attackpatterns}
\end{figure}

We used six attack triggers to evaluate the methods, as shown in Figure~\ref{fig:attackpatterns}. When more than one class is poisoned (All-to-All), we used only the grid and square triggers, positioning them at different locations based on the class index. The placement starts at the bottom right of the image and moves from right to left and bottom to top to avoid overlapping.
We ran a total of 80 experiments: for both CIFAR-10 and Tiny ImageNet, there were 24 experiments (6 triggers × 4 blending factors) for the All-to-One setting and 16 experiments (4 triggers $\times$ 4 blending factors) for the All-to-All case.  Each experiment is repeated 5 times. In each repetition $i \in\{0,\ldots, 4\}$, the random seed is set to $i$, and for the All-to-One attacks, the target class is also set to $i$. 
The $F_1$-scores and the averaged $F_1$-scores are reported as the mean over the five repetitions, along with their standard deviation. 

\subsection{Evaluation Metrics}
To evaluate the performance of poisoned instance detection, we use the $F_1$ score, defined as follows:
$$
F_1=2\times\frac{\text{Precision}\times\text{Recall}}{\text{Precision}+\text{Recall}},
\hspace*{0.4cm}
\text{Precision}=\frac{\text{TP}}{\text{TP}+\text{FP}}, \hspace*{0.4cm}
\text{Recall}=\frac{\text{TP}}{\text{TP}+\text{FN}} 
$$
where TP, TN, FP, and FN denote the number of true positives, true negatives, false positives, and false negatives, respectively.\\
\sloppy
Precision measures the proportion of correctly identified poisoned instances among all instances classified as poisoned. Recall measures the proportion of actual poisoned instances that were correctly identified by the classifier. Since it is important for a method to perform well on both metrics, the $F_1$ score, defined as the harmonic mean of precision and recall, is used to provide a balanced evaluation.

\subsection{Experimental Results}
Table~\ref{tab:badnets_detection_perf} provides the $F_1$-score of the defense methods for the All-to-One BadNets attack. SCAn and VBD achieve the best results. On average, SCAn outperforms VBD by $0.34\%$ on CIFAR-10, while VBD performs better on Tiny ImageNet, with an improvement of $1.59\%$. The average $F_1$-score across the 12 experiments indicates that VBD surpasses SCAn by $0.62\%$. Notably, AGPD exhibits low detection performance on the "Big Kitty" pattern, as this pattern covers the entire image, making it difficult for AGPD to distinguish between poisoned and clean instances.

\begin{table}[!htb]
\caption{All-to-One BadNets poisoned-instance detection $F_1$-score (\%).}
\label{tab:badnets_detection_perf}
\centering
\setlength{\tabcolsep}{9.8pt}

 \setlength\tabcolsep{4pt}
\begin{tabular}{p{0.7cm}lrrrr}
\toprule
{\bfseries Set} & 
{\bfseries Pattern} &
    \multicolumn{1}{c}{\bfseries VBD} &
    \multicolumn{1}{c}{\bfseries ABL} &
    \multicolumn{1}{c}{\bfseries AGPD} &
    \multicolumn{1}{c}{\bfseries SCAn}\\ 
\midrule
\arrayrulecolor{black!50}
 &{big kitty} &
99.87 \textpm 00.20 & 55.39 \textpm 23.40 & 79.35 \textpm 44.37 & \bftab 100.00 \textpm 00.00 \\
&{small kitty} &
99.90 \textpm 00.14 & 82.12 \textpm 10.54 & 99.62 \textpm 00.28 & \bftab 100.00 \textpm 00.00 \\
& {color grid} &
99.58 \textpm 00.15 & 90.21 \textpm 03.01 & 91.36 \textpm 12.88 & \bftab 100.00 \textpm 00.00 \\
& {green square} &
99.34 \textpm 00.25 & 84.04 \textpm 12.19 & 90.93 \textpm 11.69 & \bftab ~99.99 \textpm 00.01 \\
& {white grid} &
99.65 \textpm 00.09 & 89.66 \textpm 02.50 & 94.60 \textpm 02.64 & \bftab 100.00 \textpm 00.00 \\
& {white square} &
97.46 \textpm 00.31 & 84.30 \textpm 02.13 & 93.66 \textpm 02.96 & \bftab ~97.86 \textpm 00.12 \\
\rowcolor{gray!15!}
\cellcolor{white}
\multirow{-7}{*}{\rotatebox{90}{CIFAR-10}}& {Average} &
99.30 \textpm 00.10 & 80.95 \textpm 04.01 & 91.59 \textpm 11.65 & \bftab ~99.64 \textpm 00.02 \\
\arrayrulecolor{black!50}
 \hline
\multirow{7}{*}{\rotatebox{90}{Tiny ImageNet}} &{big kitty} &
{\bftab 99.98 \textpm 00.02 }& 81.69 \textpm 08.87 & 00.00 \textpm 00.00 & ~94.41 \textpm 12.51 \\
&{small kitty} &
99.98 \textpm 00.02 & 94.58 \textpm 07.34 & 88.90 \textpm 24.04 & \bftab 100.00 \textpm 00.00 \\
& {color grid} &
99.86 \textpm 00.12 & 96.40 \textpm 02.82 & 98.69 \textpm 00.19 & \bftab ~99.95 \textpm 00.04 \\
& {green square} &
\bftab 99.72 \textpm 00.39 & 96.77 \textpm 01.67 & 98.21 \textpm 00.66 & ~95.34 \textpm 10.35 \\
& {white grid} &
\bftab 98.70 \textpm 02.64 & 95.93 \textpm 00.84 & 98.34 \textpm 00.73 & ~97.39 \textpm 05.63 \\
& {white square} &
95.89 \textpm 02.35 & 88.66 \textpm 06.24 & 59.19 \textpm 54.03 & \bftab ~97.50 \textpm 01.26 \\
\rowcolor{gray!15!}
\cellcolor{white}& {Average} &
\bftab 99.02 \textpm 00.68 & 92.34 \textpm 02.61 & 73.89 \textpm 08.12 & ~97.43 \textpm 03.16 \\
\hline
\rowcolor{gray!30!}
\multicolumn{2}{l}{Average} &
\bftab 99.16 \textpm 00.32 & 86.65 \textpm 00.85 & 82.74 \textpm 08.96 & ~98.54 \textpm 01.58 \\
\arrayrulecolor{black}
\bottomrule
\end{tabular}
\end{table}

For the All-to-All BadNets attacks, SCAn fails on both CIFAR-10 and Tiny ImageNet as shown in Table~\ref{tab:badnets_all_to_all_detection_perf}. It appears that when at least half of the classes are poisoned SCAn is unable to detect the attack. On the contrary, VBD achieves the best $F_1$-performance, outperforming AGPD by $8.69\%$ on CIFAR-10 and ABL by $23\%$ on Tiny-ImageNet.

\begin{table}[!htbp]
\caption{All-to-All BadNets poisoned-instance detection $F_1$-score (\%).}
\label{tab:badnets_all_to_all_detection_perf}
\centering
\setlength{\tabcolsep}{9.8pt}

 \setlength\tabcolsep{4pt}
\begin{tabular}{p{0.7cm}lrrrr}
\toprule
\multicolumn{1}{c}{\bfseries Set} & 
\multicolumn{1}{c}{\bfseries Pattern} &
\multicolumn{1}{c}{\bfseries VBD} &
\multicolumn{1}{c}{\bfseries ABL} &
\multicolumn{1}{c}{\bfseries AGPD} &
\multicolumn{1}{c}{\bfseries SCAn}\\ 
\midrule
\arrayrulecolor{black!50}
 &{color grid} &
\bftab 94.37 \textpm 01.03  & 56.76 \textpm 12.58  & 68.28 \textpm 14.05  & 00.00 \textpm 00.00 \\
& {green square} &
\bftab 91.00 \textpm 02.00  & 55.36 \textpm 12.30  & 86.66 \textpm 15.02  & 00.00 \textpm 00.00 \\
& {white grid} &
\bftab 95.64 \textpm 03.54  & 58.06 \textpm 07.48  & 93.07 \textpm 09.05  & 00.00 \textpm 00.00 \\
& {white square} &
\bftab 93.51 \textpm 00.82  & 53.14 \textpm 12.94  & 91.73 \textpm 08.45  & 00.00 \textpm 00.00 \\
\rowcolor{gray!15!}
\cellcolor{white}
\multirow{-5}{*}{\rotatebox{90}{CIFAR-10}}& {Average} &
\bftab 93.63 \textpm 00.66  & 55.83 \textpm 07.63  & 84.94 \textpm 07.88  & 00.00 \textpm 00.00 \\
\arrayrulecolor{black!50}
 \hline
  & {color grid} &
\bftab 83.91 \textpm 02.75  & 50.57 \textpm 03.50  & 33.30 \textpm 00.63  & 00.00 \textpm 00.00 \\
& {green square} &
\bftab 85.00 \textpm 01.24  & 38.47 \textpm 08.08  & 34.14 \textpm 01.30  & 00.00 \textpm 00.00 \\
& {white grid} &
\bftab 52.27 \textpm 02.68  & 43.45 \textpm 06.51  & 34.34 \textpm 00.97  & 00.00 \textpm 00.00 \\
& {white square} &
\bftab 44.82 \textpm 02.46  & 41.36 \textpm 08.17  & 33.84 \textpm 00.67  & 00.00 \textpm 00.00 \\
\rowcolor{gray!15!}
\cellcolor{white}
\multirow{-5}{*}{\rotatebox{90}{\makecell{Tiny \\ ImageNet}}}
& {Average} &
\bftab 66.50 \textpm 01.91  & 43.46 \textpm 02.41  & 33.90 \textpm 00.54  & 00.00 \textpm 00.00 \\
\hline
\rowcolor{gray!30!}
\multicolumn{2}{l}{Average} &
\bftab 80.06 \textpm 01.14  & 49.65 \textpm 04.32  & 59.42 \textpm 04.14  & 00.00 \textpm 00.00 \\
\arrayrulecolor{black}
\bottomrule
\end{tabular}
\end{table}

The performance results for the Blended attacks in both All-to-One and All-to-All settings are provided in Table~\ref{tab:blended_detection_perf} and Table~\ref{tab:blended_all_to_all_detection_perf}, respectively. As we can see, in the All-to-One setting, the results are comparable to those in the All-to-One BadNets case, with SCAn and VBD leading. Nevertheless, in this case, VBD outperforms SCAn by $0.24\%$ on CIFAR-10 and by $7.37\%$ on Tiny ImageNet. The averaged $F_1$ score across the experiments on both datasets shows that VBD outperforms SCAn by $3.8\%$. For the All-to-All Blended attack case, as previously mentioned, SCAn fails, and AGPD outperforms the other defenses, with VBD finishing second. It is noteworthy that VBD struggles to detect the attack and extract the triggers when the blending factor is below $50\%$. However, when the blending factor is $50\%$, VBD performs better than AGPD.

\begin{table}[!htbp]
\caption{All-to-One Blended poisoned-instance detection mean $F_1$-score (\%) over the 6 triggers shown in Figure~\ref{fig:attackpatterns}.}
\label{tab:blended_detection_perf}
\centering
\setlength{\tabcolsep}{9.8pt}
 \setlength\tabcolsep{4pt}
\begin{tabular}{p{0.7cm}lrrrr}

\toprule
\multicolumn{1}{c}{\bfseries Set} & 
\multicolumn{1}{c}{\bfseries Blended} &
\multicolumn{1}{c}{\bfseries VBD} &
\multicolumn{1}{c}{\bfseries ABL} &
\multicolumn{1}{c}{\bfseries AGPD} &
\multicolumn{1}{c}{\bfseries SCAn}\\ 
\midrule
\arrayrulecolor{black!50}
 &{10\%} & \bftab 93.34 \textpm 01.08 & 68.60 \textpm 04.12 & 90.38 \textpm 07.27 & 92.80 \textpm 06.14 \\
& {20\%} & \bftab 96.85 \textpm 00.75 & 76.58 \textpm 01.18 & 90.22 \textpm 08.70 & 96.19 \textpm 01.10 \\
& {50\%} & 98.68 \textpm 00.54 & 84.40 \textpm 01.90 & 90.68 \textpm 09.68 & \bftab 99.14 \textpm 00.10 \\
\rowcolor{gray!15!}
\cellcolor{white}
\multirow{-4}{*}{\rotatebox{90}{ CIFAR-10}}& {Average} &
\bftab 96.29 \textpm 00.72 & 76.53 \textpm 01.99 & 90.43 \textpm 05.73 & 96.05 \textpm 02.01 \\
\arrayrulecolor{black!50}
 \hline
  &{10\%} & \bftab 83.12 \textpm 06.42 & 78.03 \textpm 03.02 & 65.02 \textpm 09.46 & 75.80 \textpm 03.19 \\
&{20\%} & \bftab 93.62 \textpm 05.91 & 87.03 \textpm 02.73 & 72.76 \textpm 19.37 & 84.58 \textpm 02.84 \\
& {50\%} & \bftab 98.05 \textpm 00.99 & 92.01 \textpm 01.67 & 79.50 \textpm 13.01 & 92.31 \textpm 03.99 \\
\rowcolor{gray!15}
\cellcolor{white}
\multirow{-4}{*}{\rotatebox{90}{\makecell{Tiny \\ ImageNet}}}
& {Average} &
\bftab 91.60 \textpm 03.80 & 85.69 \textpm 01.33 & 72.43 \textpm 11.90 & 84.23 \textpm 02.28 \\
\hline
\rowcolor{gray!30!}
\multicolumn{2}{l}{Average} &
\bftab 93.94 \textpm 01.83 & 81.11 \textpm 01.31 & 81.43 \textpm 07.04 & 90.14 \textpm 02.07 \\
\arrayrulecolor{black}
\bottomrule
\end{tabular}
\end{table}

\begin{table}[!htbp]
\caption{All-to-All Blended poisoned-instance detection mean $F_1$-score (\%) over the 4 grid and square triggers shown in Figure~\ref{fig:attackpatterns}.}
\label{tab:blended_all_to_all_detection_perf}
\centering

\setlength\tabcolsep{4pt}
\begin{tabular}{p{0.7cm}lrrrr}

\toprule
\multicolumn{1}{c}{\bfseries Set} & 
\multicolumn{1}{c}{\bfseries Blended} &
\multicolumn{1}{c}{\bfseries VBD} &
\multicolumn{1}{c}{\bfseries ABL} &
\multicolumn{1}{c}{\bfseries AGPD} &
\multicolumn{1}{c}{\bfseries SCAn}\\ 
\midrule
\arrayrulecolor{black!50}
 &{10\%} & 56.16 \textpm 03.38 & 14.08 \textpm 06.64 & \bftab 70.70 \textpm 16.82 & 00.00 \textpm 00.00 \\
& {20\%} & 72.08 \textpm 01.54 & 43.97 \textpm 05.78 & \bftab 75.29 \textpm 12.39 & 00.00 \textpm 00.00 \\
& {50\%} & \bftab 91.68 \textpm 01.27 & 53.31 \textpm 06.28 & 84.47 \textpm 04.86 & 01.40 \textpm 03.13 \\
\rowcolor{gray!15!}
\cellcolor{white}
\multirow{-4}{*}{\rotatebox{90}{ CIFAR-10}}& {Average} &
73.31 \textpm 00.80 & 37.12 \textpm 05.43 & \bftab 76.82 \textpm 05.14 & 00.47 \textpm 01.04 \\
\arrayrulecolor{black!50}
 \hline
 &{10\%} & 00.00 \textpm 00.00 & 02.84 \textpm 00.12 & \bftab 20.81 \textpm 08.53 & 00.00 \textpm 00.00 \\
&{20\%} & 09.15 \textpm 04.93 & 08.80 \textpm 03.78 & \bftab 29.04 \textpm 07.64 & 00.00 \textpm 00.00 \\
& {50\%} & \bftab 42.55 \textpm 01.09 & 37.17 \textpm 03.07 & 35.05 \textpm 00.18 & 00.00 \textpm 00.00 \\
\rowcolor{gray!15}
\cellcolor{white}
\multirow{-4}{*}{\rotatebox{90}{\makecell{Tiny \\ ImageNet}}}& {Average} &
17.23 \textpm 01.69 & 16.27 \textpm 01.94 & \bftab 28.30 \textpm 05.07 & 00.00 \textpm 00.00 \\
\hline
\rowcolor{gray!30!}
\multicolumn{2}{l}{Average} &
45.27 \textpm 00.76 & 26.69 \textpm 02.38 & \bftab 52.56 \textpm 02.43 & 00.23 \textpm 00.52 \\
\arrayrulecolor{black}
\bottomrule
\end{tabular}
\end{table}

Table~\ref{tab:summary_detection_perf} provides a summary of the performance of the tested defense methods, with the $F_1$-score averaged over all the 80 experiments we conducted. VBD ranks first on CIFAR-10 and, notably, on Tiny ImageNet. It outperforms AGPD by $3.61\%$ on CIFAR-10, and ABL by $6.25\%$ on Tiny ImageNet. SCAn is penalized by its failure in the All-to-All settings. It is worth noting that, unlike AGPD and SCAn, VBD and ABL do not rely on an auxiliary dataset. The standard deviation also shows that VBD is a stable method.

\begin{table}[!htbp]
\caption{Summary of the poisoned-instance detection $F_1$-score (\%) averaged over all the 80 experiments.}
\label{tab:summary_detection_perf}
\centering
\setlength{\tabcolsep}{9.8pt}
 \setlength\tabcolsep{4pt}
\begin{tabular}{lrrrr}
\toprule
\multicolumn{1}{c}{\bfseries Set} & 
\multicolumn{1}{c}{\bfseries VBD} &
\multicolumn{1}{c}{\bfseries ABL} &
\multicolumn{1}{c}{\bfseries AGPD} &
\multicolumn{1}{c}{\bfseries SCAn}\\ 
\midrule
\arrayrulecolor{black!50}
CIFAR-10& \bftab 89.58 \textpm 00.54 & 63.30 \textpm 03.02 & 85.97 \textpm 03.60 & 58.31 \textpm 00.60 \\
Tiny ImageNet & \bftab 67.89 \textpm 01.64 & 61.64 \textpm 00.94 & 55.56 \textpm 06.94 & 52.52 \textpm 01.36 \\
\hline
\rowcolor{gray!30!}
All & \bftab 78.74 \textpm 00.77 & 62.47 \textpm 01.28 & 70.76 \textpm 04.75 & 55.41 \textpm 00.96 \\
\arrayrulecolor{black}
\bottomrule
\end{tabular}
\end{table}

\subsection{Discussion on Explainability}
\begin{figure}[!htbp]
\centering
\begin{tabular}{c@{\hspace{0.8cm}}c@{\hspace{0.8cm}}c}
    \includegraphics[width=2.2cm]{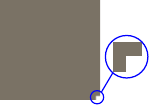} & 
    \includegraphics[width=1.5cm]{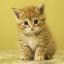} \hspace{0.6cm} &
     \includegraphics[width=2.2cm]{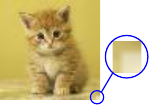} \\
     (a) trigger \hspace*{0.8cm}& (b) clean \hspace{0.55cm} & (c) malicious\hspace*{0.3cm}\\
    \\
    \includegraphics[width=2.2cm]{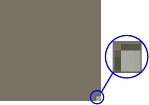} &
    \includegraphics[width=2.2cm]{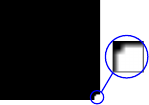} &
     \includegraphics[width=2.2cm]{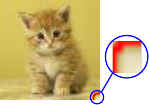}\\
     (d) extracted trigger & (e) extracted mask & (f) malicious pixels\\
\end{tabular}
\caption{Illustration of the method's explainability on a blended attack using a white square trigger (a) with a blending factor of 0.5. The pixels responsible for the malicious misclassification are outlined in red in (f). In (a) and (d), transparency is represented by the gray background.}
\label{fig:explain_ws}
\end{figure}

The proposed method extracts the salient part of the attack trigger along with its binary mask. The extracted trigger provides a quick and general visual explanation of the attack. Moreover, the binary mask can be used to isolate the pixels responsible for the malicious behavior of an instance detected as poisoned. Figure~\ref{fig:explain_ws} illustrates the explainability of the method on a blended attack on Tiny ImageNet, using a $3 \times 3$ white square trigger (shown in subfigure~\ref{fig:explain_ws}a) with a blending factor of $50\%$. The VBD method computes both the trigger and its corresponding mask, shown in subfigures~\ref{fig:explain_ws}d and~\ref{fig:explain_ws}e, respectively. The extracted pattern approximates the malicious trigger. Although the method cannot fully recover the original colors of the trigger due to the blending, the result remains sufficiently expressive to identify and explain the attack. The generated mask can then be applied to any instance detected as poisoned to isolate the malicious pixels and explain the VBD decision. The instance in subfigure~\ref{fig:explain_ws}c is detected by VBD as poisoned. The extracted mask in subfigure~\ref{fig:explain_ws}e is then used to localize the malicious pixels, as shown in subfigure~\ref{fig:explain_ws}f, where these pixels are outlined in red.

\section{Conclusion} \label{sec:conclusion}
In this article, we proposed an efficient defense algorithm against BadNets and Blended attacks. Our method does not require additional information, such as a clean dataset, and can be directly applied to a training set to detect poisoned instances. A key advantage of our approach is its explainability: it extracts the harmful part of the attack trigger, enabling experts to better understand the nature of the attack.
Moreover, our defense is effective in both the All-to-One setting, where only one class is poisoned, and the more challenging All-to-All setting, where all classes are affected. Experimental evaluations on two well-known image datasets, compared against three state-of-the-art defense methods, demonstrate the strong performance of our approach.

Our current method focuses on static-trigger attacks, which represent the simplest and most realistic backdoor attack scenarios, as they require minimal prerequisites and are thus more likely to be deployed by attackers in real-world settings. However, our approach can be extended to dynamic triggers, where the trigger is not fixed but can appear in predefined locations \cite{salem2022dynamic}. This could be achieved, for instance, by clustering the estimated poisoned set based on variance before extracting the trigger pattern.

In future work, we plan to further evaluate this extension and explore defenses against a broader range of backdoor attack strategies.

\bibliographystyle{plain}

\end{document}